\documentclass[runningheads]{llncs}
\usepackage[T1]{fontenc}
\usepackage[rgb,x11names]{xcolor}
\usepackage{amssymb}
\usepackage{amsmath}
\usepackage{subfig}
\usepackage{hyperref}
\hypersetup{colorlinks=true,citecolor=blue}
\usepackage{graphicx}
\usepackage[]{algorithm2e}
\usepackage{mathrsfs}
\usepackage{bbding}
\usepackage{rotating}
\usepackage{multirow}
\usepackage{bbm}
\usepackage{marginnote}
\usepackage{overpic}
\usepackage{colortbl}
\usepackage[labelfont=bf]{caption}
\usepackage{bbding}
\usepackage{bm}

\definecolor{mygray}{gray}{.92}

\begin{document}
\title{Unsupervised Adaptation of Polyp Segmentation Models via Coarse-to-Fine Self-Supervision}
\author{Jiexiang Wang$^1$, Chaoqi Chen$^2$\thanks{Corresponding author. J. Wang and C. Chen contributed equally to this work.}}
\institute{$^1$ByteDance, $^2$The University of Hong Kong \\
\email{wangjiexiang@bytedance.com, cqchen@gmail.com}
}
\titlerunning{Coarse-to-Fine Self-Supervision}

\maketitle  

\begin{abstract}
Unsupervised Domain Adaptation~(UDA) has attracted a surge of interest over the past decade but is difficult to be used in real-world applications. Considering the privacy-preservation issues and security concerns, in this work, we study a practical problem of Source-Free Domain Adaptation (SFDA), which eliminates the reliance on annotated source data. Current SFDA methods focus on extracting domain knowledge from the source-trained model but neglects the intrinsic structure of the target domain. Moreover, they typically utilize pseudo labels for self-training in the target domain, but suffer from the notorious error accumulation problem. To address these issues, we propose a new SFDA framework, called Region-to-Pixel Adaptation Network~(RPANet), which learns the region-level and pixel-level discriminative representations through coarse-to-fine self-supervision. The proposed RPANet consists of two modules, Foreground-aware Contrastive Learning (FCL) and Confidence-Calibrated Pseudo-Labeling (CCPL), which explicitly address the key challenges of ``how to distinguish'' and ``how to refine''. To be specific, FCL introduces a supervised contrastive learning paradigm in the region level to contrast different region centroids across different target images, which efficiently involves all pseudo labels while robust to noisy samples. CCPL designs a novel fusion strategy to reduce the overconfidence problem of pseudo labels by fusing two different target predictions without introducing any additional network modules. Extensive experiments on three cross-domain polyp segmentation tasks reveal that RPANet significantly outperforms state-of-the-art SFDA and UDA methods without access to source data, revealing the potential of SFDA in medical applications.
%and even outperforms state-of-the-art  

\keywords{Polyp Segmentation \and Source-Free Domain Adaptation \and Coarse-to-Fine Self-Supervision.}
\end{abstract}
\section{Introduction}
In clinical applications, deep learning models are typically trained on data collected from a small number of hospitals, 
but with the objective of being deployed across other hospitals to meet a broader range of needs.   
However, domain shift~\cite{quinonero2008dataset}, such as the variations of patient population and imaging conditions, hinders the deployment of well-trained models to a new hospital.
This problem has inspired a body of research on UDA~\cite{pan2009survey,guan2021domain} by explicitly mitigating the distributional shift between a labeled source domain and an unlabeled target domain~\cite{long2015learning,ganin2016domain,tzeng2017adversarial,ouyang2019data,chen2019progressive,sun2020adversarial,xia2020uncertainty,bian2020uncertainty,yu2021cross,liu2021prototypical,chen2022relation,chen2022compound}.
Despite their general efficacy for various tasks, conventional UDA techniques still have two fundamental shortcomings.
First, when encountering a new domain, UDA requires joint training of source and target data, which would be cumbersome if the number of source data is large (\emph{e.g.,} high-quality annotated data from well-known medical institutions).
Second, given the privacy-preservation issues and security concerns in medical scenarios, medical data from different clinical centers usually need to be kept locally, \emph{i.e.,} the patient data from a local hospital may not be shared to other hospitals.
In addition, in terms of memory overhead, source data mostly have larger size than source-trained models~\cite{huang2021model}, which further imposes a great challenge for real-world applications.
This motivates us to investigate a highly realistic and challenging setting called \emph{Source-Free Domain Adaptation}~(SFDA)~\cite{li2020model,liang2020we,chen2021source}, a.k.a. model adaptation, where only a trained source model and an unlabeled target dataset are available during adaptation. %process. 

Compared to mainstream UDA methods, which enable knowledge transfer with concurrent access to both source and target samples, SFDA needs to distill the domain knowledge from the fixed source model and adapt it to the target domain.
In this case, it is infeasible to leverage the prevailing feature alignment strategies in UDA, such as adversarial learning~\cite{ganin2016domain,wang2020multi}, moment matching~\cite{long2015learning}, and relation-based alignment~\cite{chen2022graphskt}, to achieve adaptation.
More importantly, we cannot explicitly measure the domain discrepancy between source and target distributions, making the adaptation process brittle to sophisticated adaptation scenarios with significant distributional shifts.
To solve this problem, most of the SFDA methods~\cite{li2020model,kundu2020universal,liang2020we,bateson2020source,liu2021source,chen2021source,liu2021adapting} is comprised of two stages, including source pre-training and target adaptation.
For example, Bateson~\emph{et al.}~\cite{bateson2020source} develop an entropy minimization term incorporated with a class-ratio prior for preventing trivial solution.
Chen~\emph{et al.}~\cite{chen2021source} propose a denoised pseudo-labeling strategy to improve the performance of self-training by collaboratively using uncertainty estimation and prototype estimation to select reliable pseudo labels.
Liu~\emph{et al.}~\cite{liu2021adapting} introduce an adaptive batch-wise normalization statistics adaptation framework, which progressively learns the target domain-specific mean and variance and enforces high-order statistics consistency with an adaptive weighting strategy. 

In spite of the fruitful progress and promising results, existing SFDA methods suffer from two key challenges.
(1) \emph{How to distinguish:} prior efforts typically learn the discriminative power from the source-trained model but ignores the self-supervised ability within the target domain for distinguishing foregrounds and backgrounds, which is crucial for exploring the intra-domain contextual and semantic structures. 
(1) \emph{How to refine:} the prevailing pseudo-labeling-based approaches~\cite{liang2020we,chen2021source,liu2021source} may be confined by false prediction, error accumulation and even trivial solution as the pseudo-labels are initialized by source-trained model and gradually refined under the absence of source supervision. %the pseudo labels can be constantly refined under the guidance of source supervision.
In a nutshell, how to simultaneously utilize domain knowledge from the source domain and mine meaningful self-supervision signals from the target domain are crucial to the success of SFDA, but remains out-of-reach for current methods.  

Remedying these issues, we propose a new SFDA framework, called Region-to-Pixel Adaptation Network~(RPANet), which unifies region-level and pixel-level representation learning in a \emph{coarse-to-fine} manner. 
The basic idea is to progressively endow the target segmentation models with the capability of distinguishing foregrounds and backgrounds via self-supervision.  
Specifically, the proposed RPANet consists of two key modules, Foreground-aware Contrastive Learning~(FCL) and Confidence-Calibrated Pseudo-Labeling (CCPL), which respectively address the challenges of ``how to distinguish'' and ``how to refine''.
FCL develops a supervised contrastive learning paradigm to learn region-level discriminative representations by contrasting different region centroids (computed by pseudo labels) across different target images, which is memory-efficient and robust to noisy pseudo labels. 
CCPL designs a novel fusion strategy to reduce the overconfidence problem of pseudo labels by fusing two different target predictions without introducing any cumbersome learning modules.
%Here, self-supervision refers to
Extensive experiments on three cross-domain polyp segmentation tasks demonstrate that our RPANet significantly outperforms state-of-the-art SFDA and UDA methods under the absence of source data.    
%existing \emph{attentive-region-guided alignment} and \emph{object-relation-embedded alignment} approaches are still brittle to 
%sophisticated adaptation scenarios with large domain discrepancies as the topological relations among foreground regions remain unexplored.

%Very recently, several SFDA methods have been proposed to  
\section{Region-to-Pixel Adaptation Network}

\begin{figure*}[!t]
	\centering
	\includegraphics[width=1\textwidth]{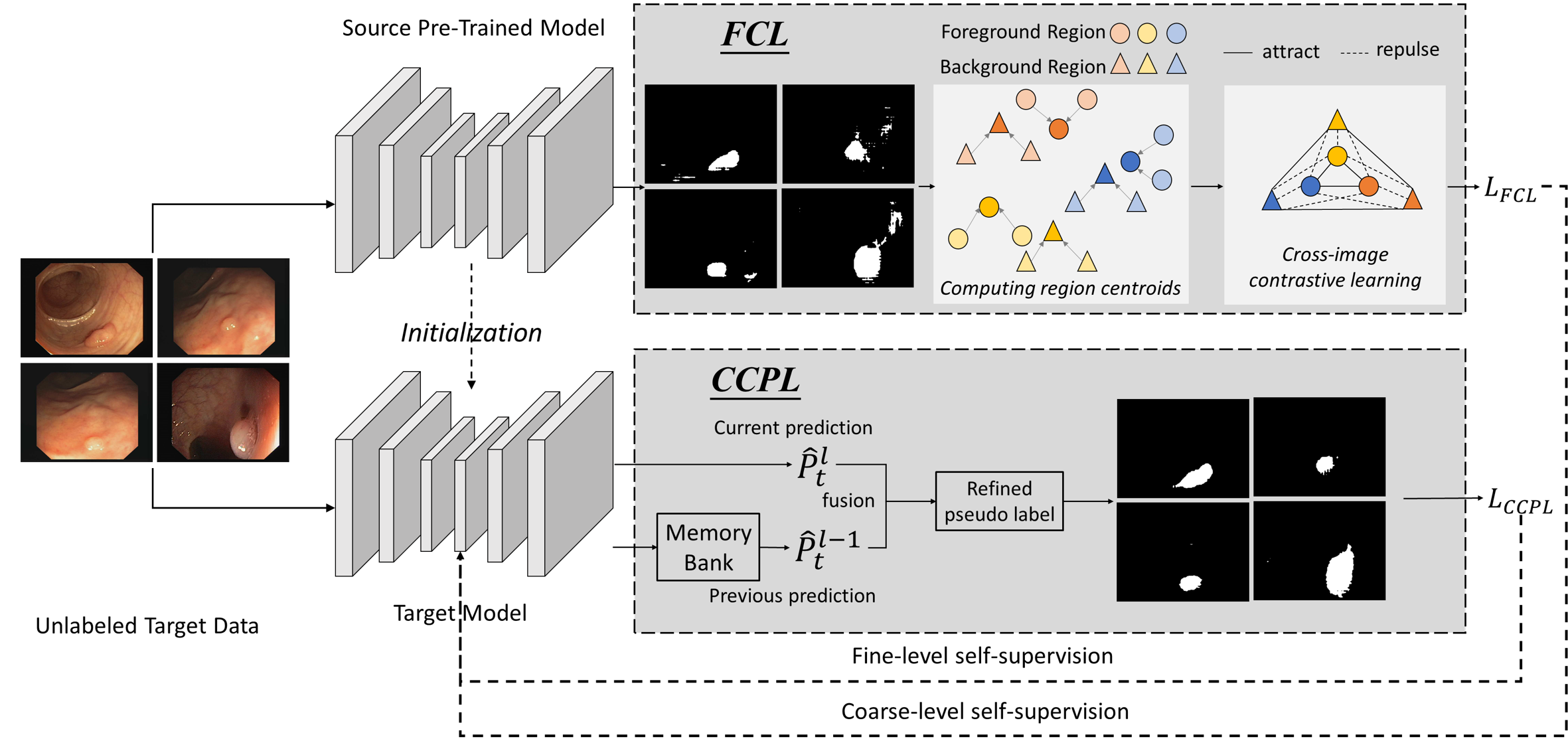}
	\caption{Overview of our Region-to-Pixel Adaptation Network~(RPANet), which includes a pre-trained source model, a target model, and two model adaptation modules (\emph{i.e.,} FCL and CCPL). These two model adaptation modules work in a coarse-to-fine manner to learn discriminative representations in the two levels.}
	%\vspace{-0.5cm}
	\label{fig1}
\end{figure*}

In this paper, we investigate the problem of SFDA, where we only have access to a pre-trained source model $\bm{f}_s$ and an unlabeled target dataset 
$\mathcal{D}_t=\{x_t|x_t\in{\mathbb{R}}^{H \times W \times 3}\}$. 
$\bm{f}_s$ was trained on labeled source data $\mathcal{D}_s$. 
The source and target data do not follow the IID assumption.
The objective of SFDA is to learn a segmentation model that performs well on the target domain.
Figure~\ref{fig1} illustrates an overview of RPANet, which consists of two components, namely, Foreground-aware Contrastive Learning (FCL) and Confidence-Calibrated Pseudo-Labeling (CCPL).
In particular,
FCL and CCPL are complementary to each other, \emph{i.e.,}
FCL enhances the robustness of CCPL by providing region-level discriminative representations as an initialization, 
while CCPL refines the target pseudo labels to mitigate the bias introduced by FCL.   

\subsection{Foreground-aware Contrastive Learning}
%motivation + step
Contrastive learning~\cite{he2020momentum,khosla2020supervised} has achieved compelling results in self-supervised representation learning by making the representations of the positive pair close while keep negative pairs apart. 
In view of the dense-prediction property of segmentation tasks, some of the prior works~\cite{chaitanya2020contrastive,wang2021dense} utilize contrastive learning as a pre-training step. 
They typically conduct this learning process in the image level (\emph{i.e.,} the augmentations of original image are regarded as positive samples and other images from the dataset are regarded as negative samples) while neglecting the holistic context of the entire dataset, \emph{i.e.,} the relations among different real images. 
%rather than generated ones. 
The rationale is that these approaches rely on image augmentation techniques to make contrast under the absence of ground-truth annotations. 
%Since no annotation is available for target model adaptation,
%or directly conduct pixel-to-pixel contrastive learning in a supervised manner. 
%Unfortunately, since no annotation is available for target model adaptation, 
In SFDA, we also have the same dilemma.
Instead of following the conventional wisdom, we make use of pseudo labels~\cite{lee2013pseudo} predicted from the source pre-trained model to perform contrastive learning in a \emph{supervised} way.

%given that pseudo labels generated from source-trained model would be noisy.
%More importantly, as opposite to prevailing instance-wise contrastive learning paradigms~\cite{zhao2021makes}, we propose to xxx grounded on the observation that 

Then, an intuitive solution is to utilize pseudo labels to perform pixel-level contrastive learning.
However, 1) the pseudo labels may be noisy, and 2) the pixel-level annotations would bring in huge memory overhead for constructing the memory bank.
To solve these issues, we propose the FCL module to contrastively learn region-level discriminative representations in a fully supervised way. 
Here, we use the region centroid to stand for the very region to reduce computational cost. 
In addition, when computing the centroid, we consider the prediction confidence of different pixels to dynamically assign weights to them.

Formally, the source-trained model $\bm{f}_s$ takes a target image $x_t$ as input and produces the prediction map $P_t=\bm{f}_s(x_t)$, and its corresponding feature map is denoted as $F_t\in\mathbb{R}^{H \times W \times K}$. 
The entropy map $I_t$ is defined as,
\begin{equation}
	I_t^{(h,w)} = H(P_t),
\end{equation}
where $H(\cdot)$ is the entropy function. Then, the region centroid of class $k$ is formulated as,
\begin{equation}
	m_k = \frac{\sum_{h,w}F_t\mathbbm{1}[P_t^{(h,w)}=k]\cdot(1-I_t^{(h,w)})}{\sum_{h,w}\mathbbm{1}[P_t^{(h,w)}=k]}
\end{equation}
where $\mathbbm{1}(\cdot)$ is an indicator function, and each pixel is re-weighted by its prediction confidence when computing the centroid. %which is 1 if $a$ is true and 0 otherwise.
%The generator $G_{inter}$ takes a target image $X_t$ as input to generate $P_t$ and the entropy map $I_t$. On this basis, we adopt a simple yet effective way for ranking by using:
%\begin{equation}
%	R(X_t) = \frac{1}{HW} \sum_{h,w}I_t^{(h,w)}, 
%\end{equation}
%which is the mean value of entropy map $I_t$. 
After that, the region-level contrastive loss regarding category $k$ can be formulated as,
%After constructing the region centers of all categories for each image, region-aware contrastive loss for a region center
%from class $i$ can be directly defined as,
\begin{equation}
	%\small
	\label{eq:FCL}
	\!\!\!\!\mathcal{L}^{\text{FCL}}_k\!=\!-\frac{1}{|\mathcal{M}_k|}\sum_{m^+\in{M_k}}\log\frac{\exp(\bm{m}\!\cdot\!\bm{m}^+/\tau)}{\exp(\bm{m}\!\cdot\!\bm{m}^+/\tau)
		\!+\!\sum_{\bm{m}^-\in \mathcal{N}_k}\exp(\bm{m}\!\cdot\!\bm{m}^-/\tau)}\!\!
\end{equation}
where $\mathcal{M}_k$ and $\mathcal{N}_k$ stand for the collections of the positive and negative samples, for $\bm{m}$. $m^+$ and $m^-$ stand for the certain positive and negative samples, respectively.
%where 
%, instead of pushing all the pixel embeddings into memory banks, we just push several region centers from different categories into the banks.
%advantages
As its core, FCL leverage the ``learning to compare'' ability of contrastive learning to make a clear distinction between foreground and background regions as well as model the cross-image foreground/background relations.
%The core idea of FCL is ``learning to compare'': given an anchor point, 
%exploits the intrinsic contextual dependencies within target domain to induce contrastive learning in the region level, 
%thereby coarsely highlighting the foreground regions.
By doing so, regions that are more likely to be the foregrounds are coarsely highlighted.

\subsection{Confidence-Calibrated Pseudo-Labeling}
Pseudo-labeling~\cite{lee2013pseudo} %(a.k.a. self-training)
has been proved to be a simple yet effective approach for SFDA~\cite{liang2020we,chen2021source,liu2021source}.
Most of state-of-the-art pseudo-labeling methods prioritize their focus on designing sampling strategies to select pseudo labels with high prediction confidence and fit task-specific properties.
%dynamic re-weighting strategies to adaptively adjust the confidence of pseudo labels, 
However, they ignore the structured dependencies of different pixels inside a single image, which is prone to result in overconfident predictions regarding some local regions. 
%The key point of PLST is to design a pseudo label weighting strategy.
%In that way, these methods capture the contextual relationship between images, 
More importantly, in some medical applications, where the foreground and background are highly entangled, these methods may be prohibitively difficult to select trustworthy pseudo labels. 
For example, in colonoscopy images, polyps and normal tissues are visually similar and have low contrast, greatly impeding the correct assignment of pseudo labels.
To address this problem, the proposed CCPL calibrates the outputs soft pseudo labels in the pixel level, which integrates the output of previous and current predictions via a simple yet effective fusion strategy.

%incorporating the structural information and producing soft pseudo labels in the pixel level.  
 
 %by devising intra-image and inter-image calibration mechanisms. 
%These two mechanisms are orthogonal yet complementary to each other.

%\textbf{Intra-image calibration.}
%\textbf{Inter-image calibration.}
%Note that intra-image calibration ignore the connection of different images, which may obtain a limit performance. Here we explore the contextual relationship between different augmented images.
%Specifically, we resort to ensemble multiple 
Given the output of the target segmentation network $\hat{P}_t=\bm{f}_t(x_t)$ (pre-softmax logits), 
the target pseudo-label is updated as follows,
\begin{equation}
    \hat{P}_t^l = \alpha\cdot\text{softmax}(\frac{\hat{P}_t^l}{\phi(\hat{P}_t^l,\hat{P}_t^{l-1})})+(1-\alpha)\cdot\text{softmax}(\frac{\hat{P}_t^{l-1}}{\phi(\hat{P}_t^l,\hat{P}_t^{l-1})})
\end{equation}
where $\hat{P}_t^{l-1}$ is the pre-softmax logits of $(l-1)$-th times, $l$ denotes the iteration times,
and $\alpha$ (ranged from 0 to 1) is a modulated factor. In practice, we set $\alpha=0.5$ in all experiments.
%are the segmentation results corresponding to weak and strong augmented images, which 
%could have various degrees of overconfident. 
%To facilitate a fair comparison, we use standard weak augmentation~\cite{chen2018encoder} in experiments.
The function $\phi(a,b) = \sqrt{\sum_i^{|a|}{(a_i^2+b_i^2)}}$ is a normalization term, which could smooth the pseudo label probability distribution to avoid overconfident predictions (dominate pixels). 
 
After obtaining reliable pseudo labels, %we directly conduct dot multiplication on both probabilities to get the final soft pseudo label $P_{t}=P_t^{intra}{\cdot}P_t^{inter}$ and its one-hot label $Y_t^{pl}$.
the segmentation network is optimized in a supervised way by minimizing the cross entropy loss,
\begin{equation}
    \mathcal{L}_{\rm CCPL}=-\sum_{h,w}\sum_{c}\hat{Y}_t^{(h,w,c)}\log(P_t^{(h,w,c)})
\end{equation}
where $\hat{Y}_t^{(h,w,c)}$ is the soft pseudo labels.

\subsection{Objective Function}
First of all, we utilize the weights of source pre-trained model $\bm{f}_s$ for the initialization of target model.
The overall training objective of RPANet, which includes FCL and CCPL, can be formulated as follow,
\begin{equation}\label{eq:overall}
	\mathcal{L}_{\rm RPANet}=\beta\sum\limits_k\mathcal{L}^{\text{FCL}}_k + \gamma\mathcal{L}_{\rm CCPL}
\end{equation}
where $\beta$ and $\gamma$ are two trade-off parameters.
%\newpage
\section{Experiments} 
\subsection{Dataset}
We extensively evaluate the proposed method RPANet on the polyp segmentation tasks with three public datasets and an in-house dataset.
(1) ClinicDB~\cite{bernal2015wm} contains 612 Standard Definition (SD) frames from 31 sequences.
(2) ETIS-LARIB~\cite{silva2014toward} contains 196 High Definition (HD) frames from 34 sequences.
(3) Kvasir-SEG~\cite{jha2020kvasir} contains 1000 polyp frames with various resolutions.
(4) In-house dataset is collected from a local hospital and contains 5175 frames with polyp. The annotations are sketched by two experienced gastroenterologists.
In experiments, considering the number of annotated data in different datasets, we use in-house dataset as the source domain, and the other three public datasets as the target domain respectively.
\subsection{Implementation Details and Evaluation Metrics}
We adopt DeepLab-v2~\cite{chen2017deeplab} as the segmentation model and ResNet101~\cite{he2016deep} pre-trained on ImageNet as the backbone network. 
In the training phase, we use Stochastic Gradient Descent (SGD) with momentum 0.9 as the optimizer.
%and weight decay 1e-4 . 
The learning rate is set as $2.5 \times 10^{-4}$, which is decayed by a polynomial annealing policy~\cite{chen2017deeplab}. 
The batch size is 4 and we train the models for 20 epochs. 
Following prior works, standard data augmentation techniques are utilized in experiments.
For Eq.~(\ref{eq:FCL}), we set the temperature $\tau$ as 0.1.
%We set $\tau=$ in Eq.~(\ref{eq:FCL}).
For Eq.~(\ref{eq:overall}), we set $\beta=1$ and $\gamma=1$  in all experiments.
%to avoid over-fitting in the training procedure.
All experiments are conducted on 4 NVIDIA Tesla V100 GPUs with PyTorch deep learning framework. 
Following~\cite{fan2020pranet}, we use 6 metrics to quantitatively evaluate the superiority of our method, including mean Dice, mean IoU, weighted Dice metric $F_{\beta}^{w}$, structure similarity measure $S_{\alpha}$, enhanced-alignment metric $E_{\phi}^{max}$, and MAE metric.

\begin{table*}[!t]
	\centering
	%\scriptsize
	%\renewcommand{\arraystretch}{1.1}
	\setlength\tabcolsep{5pt}
	\caption{Quantitative results of different adaptation methods on ClinicDB~\cite{bernal2015wm}, ETIS-LARIB~\cite{silva2014toward}, and Kvasir-SEG~\cite{jha2020kvasir} datasets.}
	\label{tab1}
        \scalebox{0.93}{
	\begin{tabular}{c|r||cccccccc}
		\hline
		\rowcolor{mygray}
		&Methods & mean Dice & mean IoU  &  $F_\beta^w$ & $S_{\alpha}$&$E_\phi^{max}$ & MAE \\
		\hline
		\multirow{7}{*}{\begin{sideways}ClinicDB\end{sideways}} & 
		w/o adaptation  &  67.3 & 55.5 & 63.9 & 76.7 & 82.8 & 0.047\\
		&BDL~\cite{li2019bidirectional} &  76.8 & 67.3 & 73.8 & 83.3 & 87.1 & 0.037\\
		&FDA~\cite{yang2020fda} & 78.6 & 71.1 & 78.7 & 85.6 & 87.6 & 0.027\\
		&HCL\cite{huang2021model} & 74.1 & 64.5 & 74.5 & 82.4 & 85.2 & 0.034\\
		&DPL~\cite{chen2021source}& 76.3 & 66.9 & 75.3 & 83.7 & 87.0 & 0.034\\
		&RPANet w/o FCL & 75.7 & 66.1 & 74.9 & 83.2 & 86.7 & 0.035\\
		&RPANet w/o CCPL & 73.2 & 62.8 & 70.6 & 80.7 & 86.4 & 0.041\\
		&\textbf{RPANet} & \textbf{80.0} & \textbf{71.9} & \textbf{80.2} & \textbf{86.4} & \textbf{89.2} & \textbf{0.029}\\
		\hline
		\hline
		\multirow{7}{*}{\begin{sideways}ETIS-LARIB\end{sideways}} &
		w/o adaptation  &  49.6 & 41.4 & 47.2 & 70.9 & 70.6 & 0.026\\
		&BDL~\cite{li2019bidirectional} &  59.3 & 51.8 & 57.1 &76.3 &74.8 &0.019\\
		&FDA~\cite{yang2020fda} & 61.5 & 53.5 & 59.7 & 77.3 & 77.0 & 0.017\\
		&HCL\cite{huang2021model} & 57.3 & 49.5 & 55.4 & 75.1 & 73.8 & 0.024\\
		&DPL~\cite{chen2021source}& 58.8 & 51.4 & 56.6 & 76.1 & 74.5 & 0.020\\
		&RPANet w/o FCL & 59.1 & 51.6 & 56.9 & 76.3 & 74.6 & 0.019\\
		&RPANet w/o CCPL & 59.9 & 52.2 & 58.1 & 76.7 & 75.5 & 0.017\\
		&\textbf{RPANet} & \textbf{63.2} & \textbf{55.2} & \textbf{61.3} & \textbf{78.3} & \textbf{77.9} & \textbf{0.016}\\
		\hline
		\hline
		\multirow{7}{*}{\begin{sideways}Kvasir-SEG\end{sideways}} &
		w/o adaptation  &  69.0 & 57.5 & 62.4 & 74.2 & 79.6 & 0.109 \\
		&BDL~\cite{li2019bidirectional} &  84.2 & 76.0 & 82.3 & 86.7 & 91.1 & 0.044\\
		&FDA~\cite{yang2020fda} & 84.6 & 76.5 & 83.0 & 87.0 & 91.4 & 0.042\\
		&HCL\cite{huang2021model} & 84.0 & 75.8 & 82.4 & 86.6 & 91.0 & 0.043\\
		&DPL~\cite{chen2021source}& 80.4 & 71.0 & 79.0 & 84.1 & 88.5 & 0.050\\
		&RPANet w/o FCL & 80.5 & 71.7 & 76.6 & 83.5 & 88.0 & 0.062\\
		&RPANet w/o CCPL & 79.2 & 69.7 & 75.3 & 82.5 & 87.4 & 0.063\\
		&\textbf{RPANet} & \textbf{85.8} & \textbf{78.2} & \textbf{84.4} & \textbf{87.9} & \textbf{92.1} & \textbf{0.039}\\
		\hline
		%\vspace{-10pt}
	\end{tabular}}
\end{table*}

\subsection{Comparisons with State-of-the-Arts}
\textbf{State-of-the-Arts.}
We extensively compare the proposed RPANet with state-of-the-art domain adaptive semantic segmentation methods, including Bidirectional Learning~\textbf{(BDL)}~\cite{li2019bidirectional}, Fourier Domain Adaptation~\textbf{(FDA)}~\cite{yang2020fda}, Historical Contrastive Learning~\textbf{(HCL)}~\cite{huang2021model}, and Denoised Pseudo-Labeling~\textbf{(DPL)}~\cite{chen2021source}.
In our polyp segmentation experiments, we use the provided source codes to implement these baseline methods.
Note that BDL and FDA are conventional UDA methods, while HCL and DPL are SFDA methods. 
``W/o adaptation'' denotes that the model is trained on the source data and directly applied to the target data without any adaptation. 
%All these methods are based on PLST framework to alleviate performance discrepancy in the target domain. 
%Note that all these methods are implemented on the same network(DeepLab-V2) for fair comparison. 

\begin{figure*}[thb]
	\centering
	\includegraphics[width=1\textwidth]{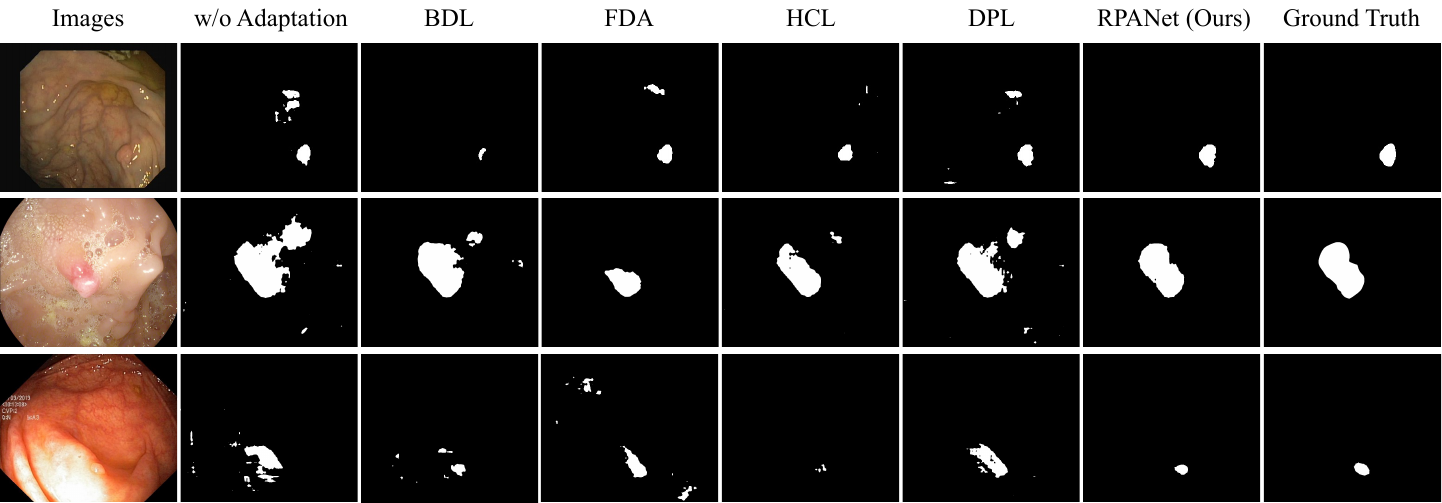}
	%\vspace{-0.2cm}
	\caption{Qualitative evaluation of different adaptation methods on target domains. From top to bottom: ClinicDB, ETIS-LARIB, Kvasir-SEG.}
	\vspace{-0.5cm}
	\label{fig:result1}
\end{figure*}

\begin{figure*}[htb]
	\centering
	\includegraphics[width=1\textwidth]{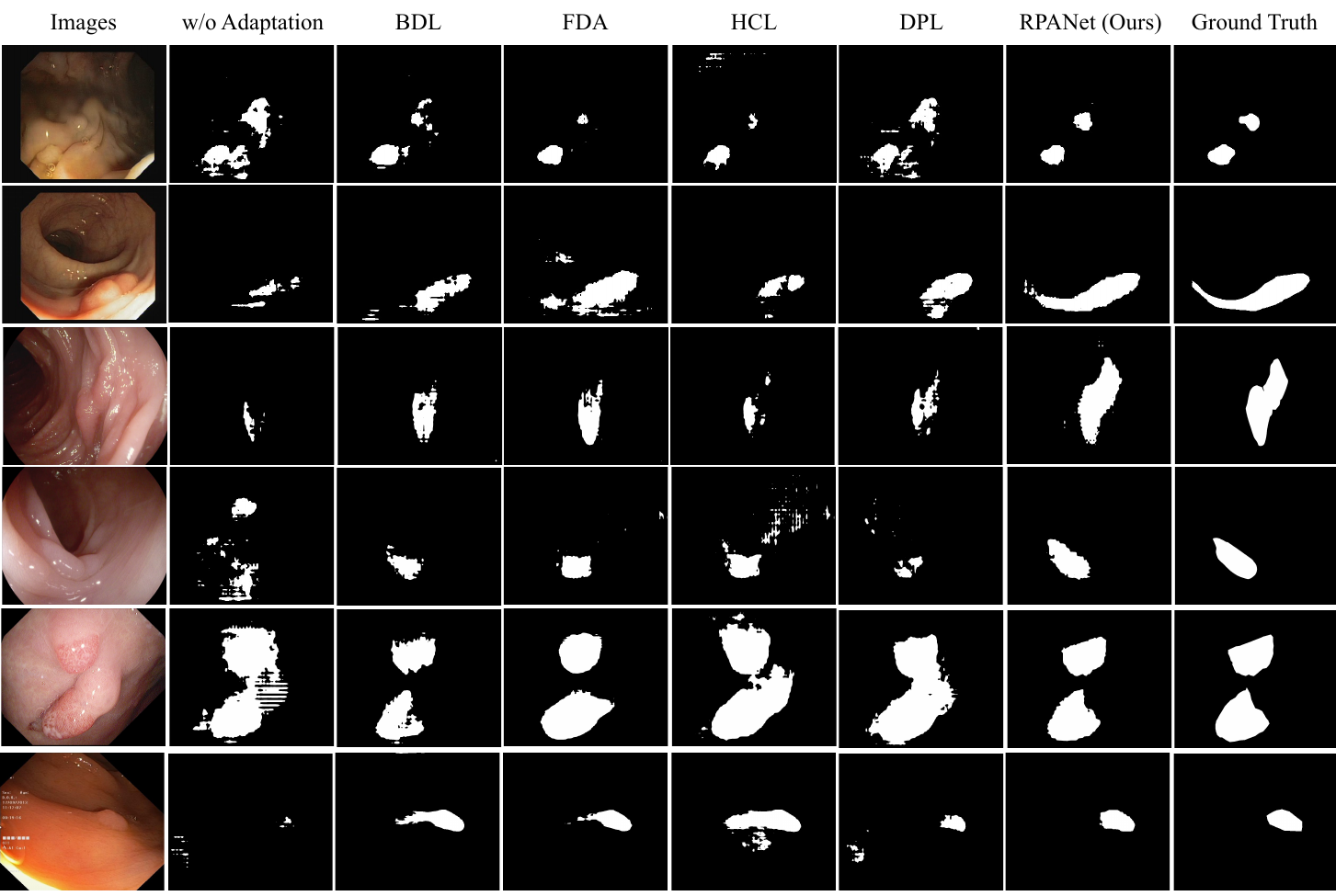}
	%\vspace{-0.2cm}
	\caption{Qualitative evaluation of different adaptation methods on target domains. From top to bottom: ClinicDB, ETIS-LARIB, Kvasir-SEG.}
	%\vspace{-0.5cm}
	\label{fig:result2}
\end{figure*}

%2)Segmentation Results:
Table~\ref{tab1} reports the performance comparisons of different adaptation methods on three target domains. 
The proposed RPANet consistently outperforms all compared methods on all domain adaptation tasks. 
RPANet significantly promotes the w/o adaptation result from 62.0\% to 76.3\% on average in terms of mean Dice, which brings about 14.3\% improvement. 
In particular, the performance of HCL and DPL cannot be on par with the conventional UDA methods (BDL and FDA), revealing the importance and difficulty of adapting polyps segmentation models without access to source data. 
By contrast, RPANet is almost systematically better than the state-of-the-art UDA methods, highlighting the contributions of our coarse-to-fine self-supervision scheme.
%HCL and DPL are source-free based method, the former xxx, the latter xxx. 
%Both two methods do not consider xxx, so they cannot solve the xxx problem. 
%BDL and FDA use source data for adaptation to obtain more satisfactory result, but both of them can't work without touching source data. Moreover, the methods above has a common weak point that xxx. 
To qualitatively illustrate the superiority of our method, we show the examples of polyps segmentation results in Figure~\ref{fig:result1} and Figure~\ref{fig:result2}. %qualitatively  the examples of polyps segmentation results on three target domains. 
We can observe that our RPANet is capable of precisely segmenting polyps, reducing ambiguous predictions, and refining the boundaries between lesions and normal tissues. 
The justification is that RPANet structurally regularizes the outputs via FCL and CCPL, thereby solving the problem of incomplete and noisy predictions.  

\begin{center}
	\begin{table}[t]
		\caption{Further empirical analysis regarding the different design choices of RPANet. Mean Dice is reported on the target domains.}
		\setlength\tabcolsep{6pt}
		\centering
		\label{tab:integration}
		\begin{tabular}{cccc}
			\hline
			\rowcolor{mygray}
			Method & ClinicDB & ETIS-LARIB & Kvasir-SEG \\
			\hline
			RPANet &  80.0 & 63.2 & 85.8\\
			\hline
			\multicolumn{4}{c}{Integration with modern SFDA methods} \\
			\hline
			FCL + HCL~\cite{huang2021model} &  79.3 & 59.1 & 85.6\\ 
			FCL + DPL~\cite{chen2021source} &  78.5 & 60.1 & 84.1\\
			CCPL + HCL~\cite{huang2021model} &  78.6 & 58.8 & 84.9\\
			CCPL + DPL~\cite{chen2021source} &  77.4 & 60.3 & 83.6\\
			\hline
			\multicolumn{4}{c}{Ablation of FCL and CCPL} \\
			\hline
			%FCL w/  &  \\
			FCL w/o entropy regularization & 78.9 & 62.6 & 85.2\\
			%CCPL w/ & \\
			CCPL w/o confidence calibration & 78.3 & 61.8 & 84.1\\
			\hline
			\vspace{-0.5cm}
		\end{tabular}
	\end{table}
\end{center}

%\begin{figure*}[thb]
%	\centering
%	\includegraphics[width=1\textwidth]{entropymap.pdf}
%	%\vspace{-0.2cm}
%	\caption{Qualitative evaluation of different adaptation methods on target domains. From top to bottom: ClinicDB, ETIS-LARIB, Kvasir-SEG.}
%	%\vspace{-0.5cm}
%	\label{fig2}
%\end{figure*}

\vspace{-1cm}
\subsection{Further Empirical Analysis}
\textbf{Ablation Study.}
To investigate the individual effect of each component (\emph{i.e.,} FCL and CCPL) in our proposed RPANet, 
we conduct ablation experiments on three domain adaptive polyp segmentation benchmarks. 
The results are also presented in Table~\ref{tab1}. 
From the table, we can observe that both FCL and CCPL are reasonably designed, and when one of them are removed, the final performance would drop accordingly. 

\noindent
\textbf{Design Choice.} 
We further explore the alternative design choices for each component to better understand the essence of RPANet. (1) Integration with modern SFDA methods. (2) Ablation of FCL and CCPL.
The results are demonstrated in Table~\ref{tab:integration}.
We can see that both FCL and CCPL substantially improve state-of-the-art SFDA methods by providing more accurate self-supervision signals. In addition, when the elaborately devised regularization and calibration terms are removed, the performance will be affected to some extent.

\begin{figure*}[thb]
	\centering
	\includegraphics[width=1\textwidth]{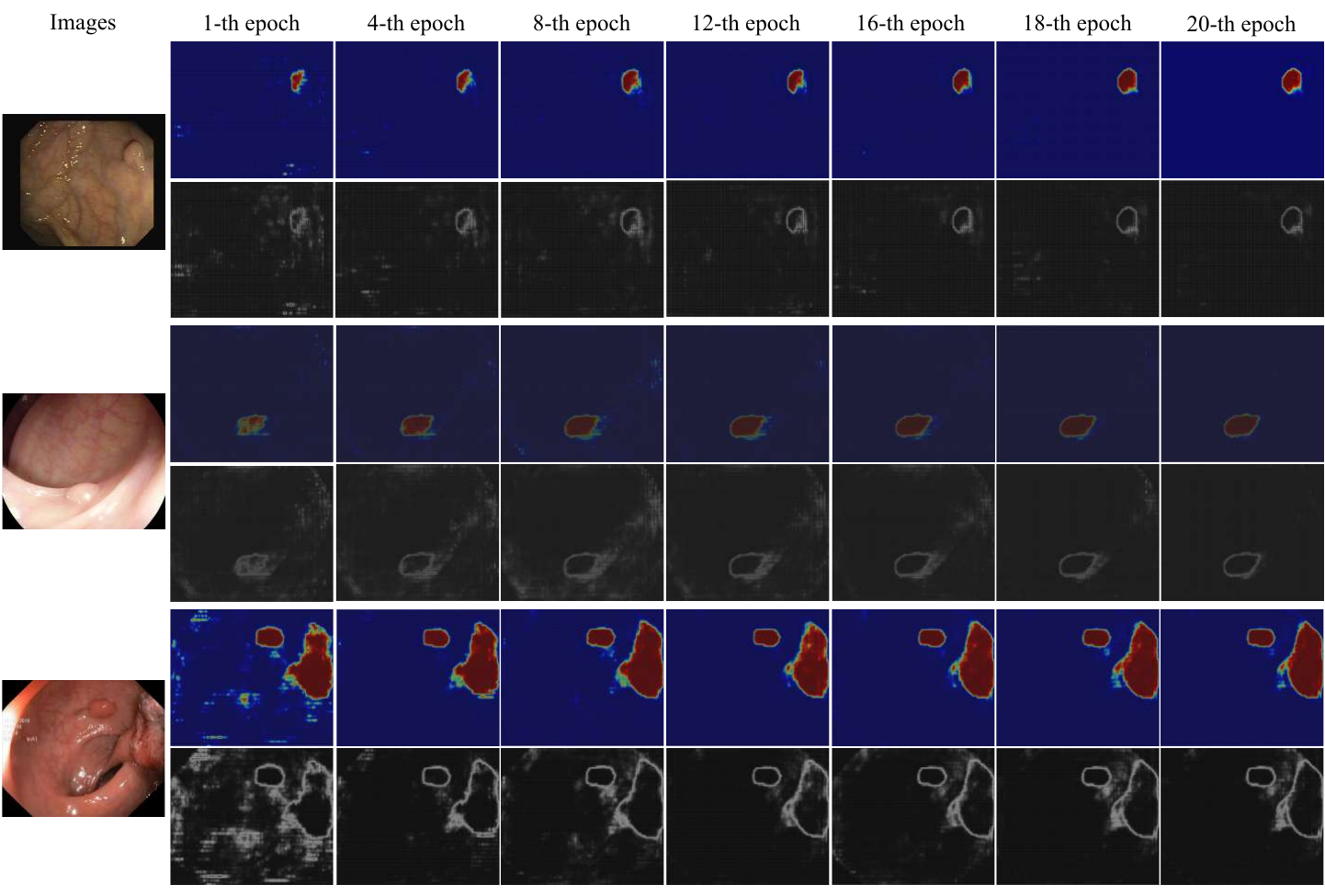}
	%\vspace{-0.2cm}
	\caption{Visualization results on target images as training proceeds. From top to bottom: ClinicDB, ETIS-LARIB, Kvasir-SEG. In every two rows, \textbf{upper:} heat map, \textbf{lower:} entropy map. Noting that the brighter the color is, the larger the corresponding value is.}
	%\vspace{-0.5cm}
	\label{fig:visualization}
\end{figure*}

\noindent
\textbf{Visualization.} 
We visualize the heat map and entropy map on some challenging target images (\emph{i.e.,} the boundaries between the polyp and its surrounding tissues are very ambiguous) as training proceeds in Figure~\ref{fig:visualization}. From the figure, we found that the shapes and boundaries are progressively refined as the training epoch increases, which clearly demonstrates the effectiveness of our coarse-to-fine refinement paradigms.
In particular, we can observe that our RPANet achieves competitive results even with very limited training times, such as 8 epochs.

\begin{figure*}[!h]
	\centering
	%\small
	\setlength\tabcolsep{1mm}
	\renewcommand\arraystretch{0.1}
	\begin{tabular}{ccc}
		\includegraphics[width=0.33\linewidth]{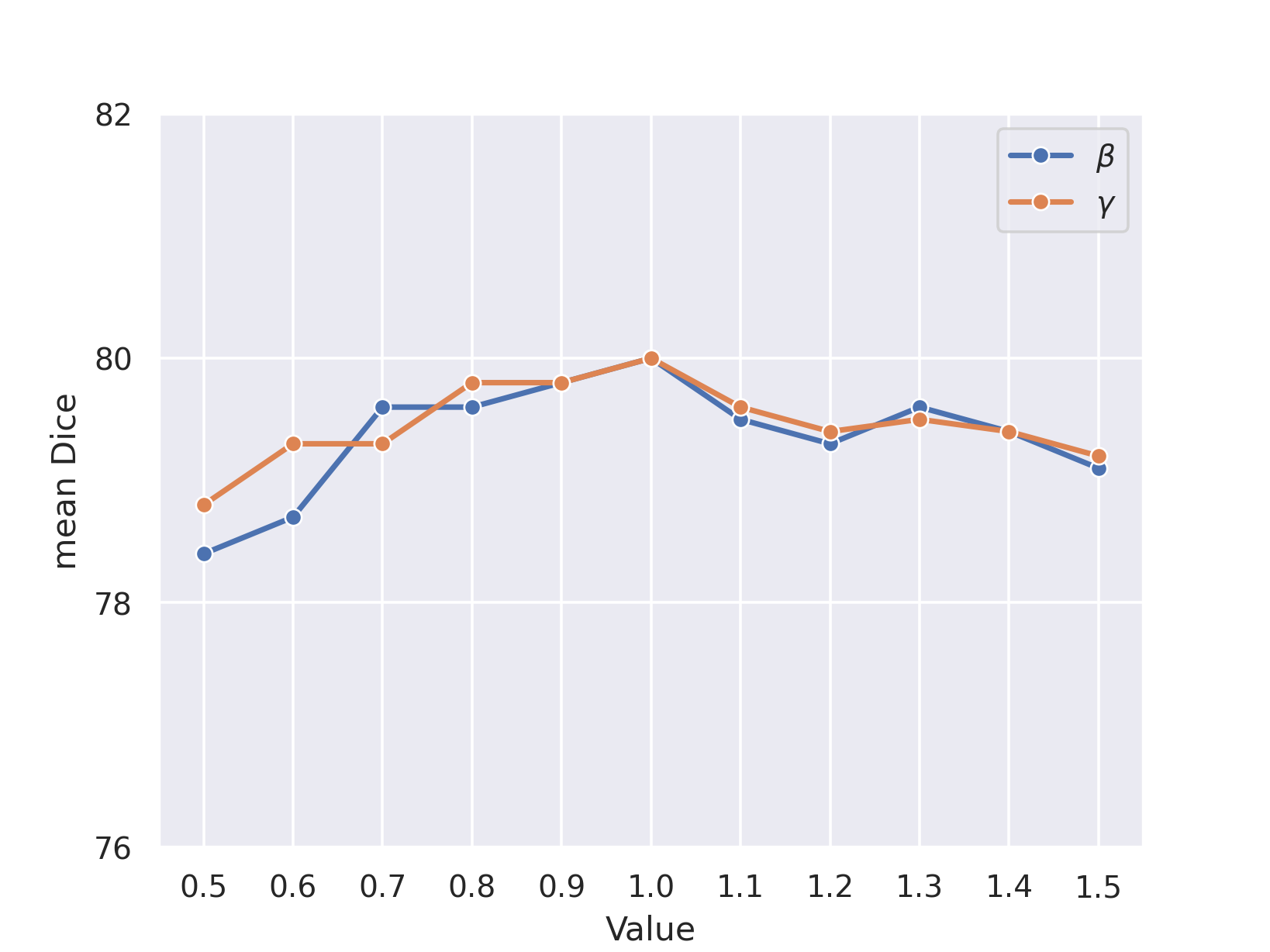} &
		\includegraphics[width=0.33\linewidth]{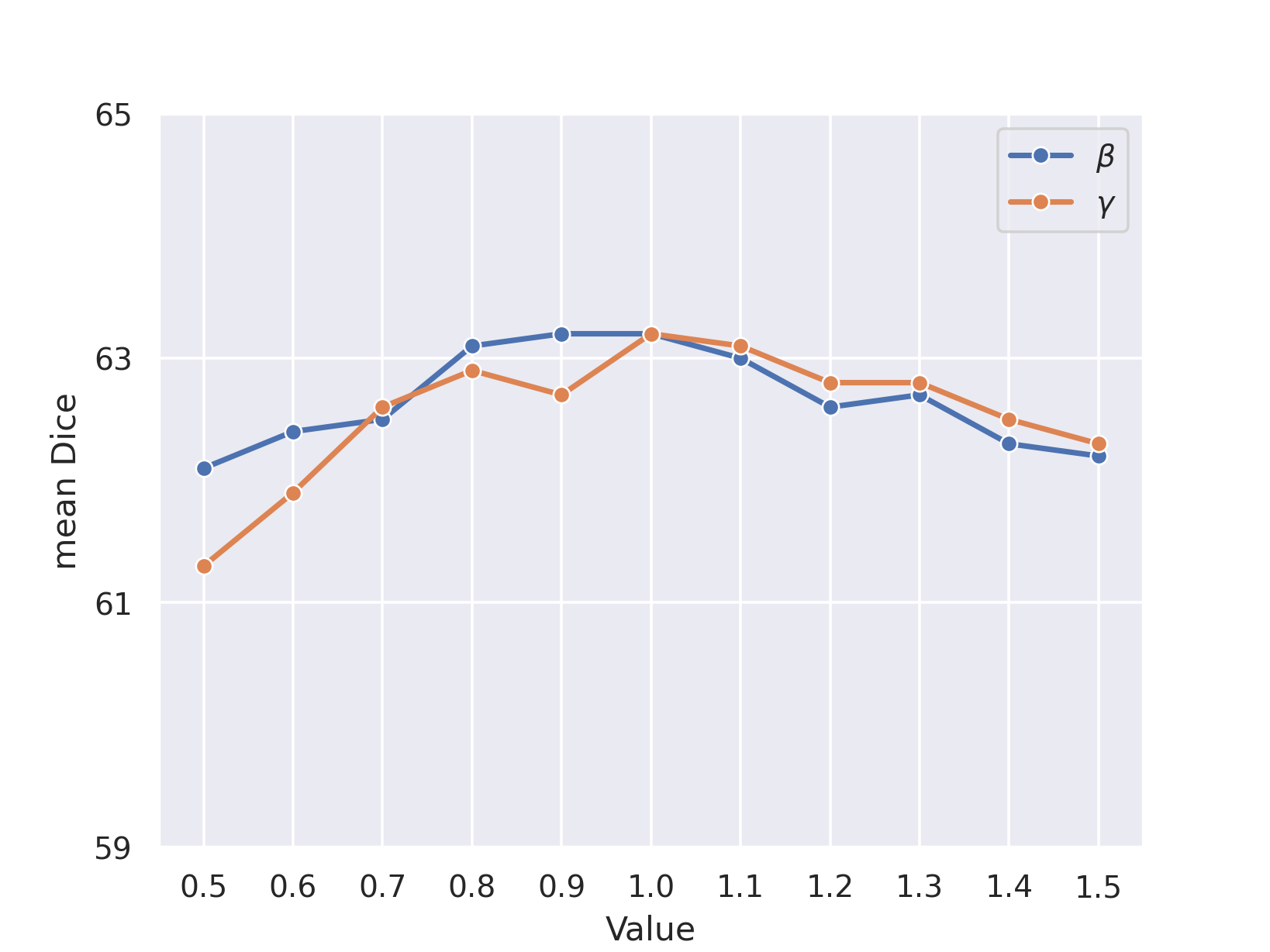} & 
		\includegraphics[width=0.33\linewidth]{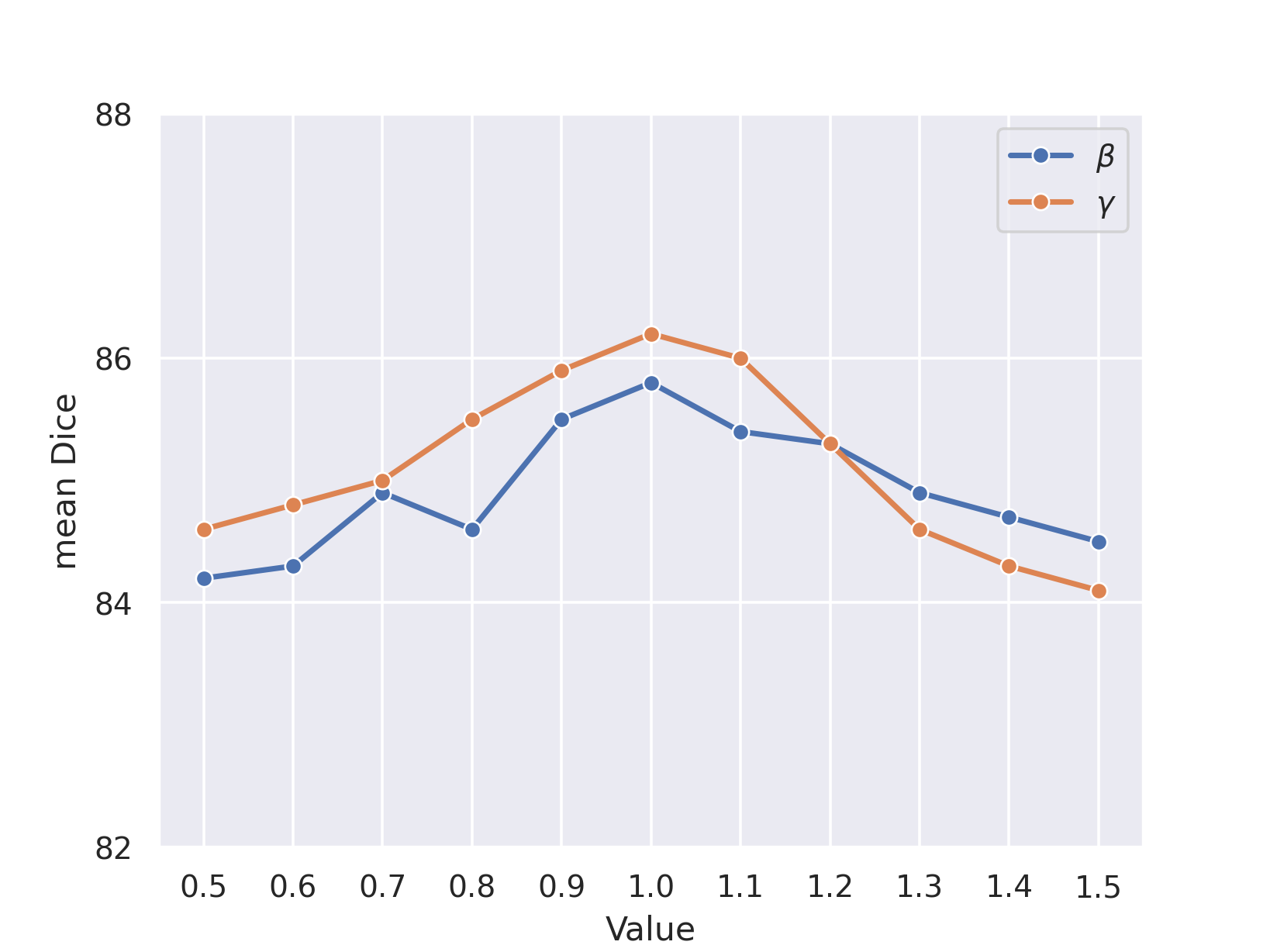} \\
		%~\\
		%~\\
        %\vspace{10pt}
	(a) ClinicDB  & (b) ETIS-LARIB & (c) Kvasir-SEG  \\
	\end{tabular}
	\caption{Sensitivity analysis regarding hyper-parameters $\beta$ and $\gamma$.}
	\label{fig:hyper}
	%\vspace{10pt}
\end{figure*}

\noindent
\textbf{Parameter Sensitivity.}
We provide the sensitivity analysis regarding hyper-parameters $\beta$ and $\gamma$ in Figure~\ref{fig:hyper}. As can be seen, the performance of our RPANet is insensitive to the variations of the value of $\beta$ and $\gamma$ on different benchmark datasets, revealing the robustness of the proposed modules. For example, changing $\beta$ in the range [0.5, 1.5] only incurs small performance variations \emph{i.e.,} 1.6\%, 1.1\%, and 1.6\% in ClinicDB, ETIS-LARIB, and Kvasir-SEG respectively.

% \begin{figure*}[thb]
% 	\centering
% 	\includegraphics[width=1\textwidth]{hyper.pdf}
% 	%\vspace{-0.2cm}
% 	\caption{Sensitivity analysis regarding hyper-parameters $\beta$ and $\gamma$. From left to right: ClinicDB, ETIS-LARIB, Kvasir-SEG.}
% 	%\vspace{-0.5cm}
% 	\label{fig:hyper}
% \end{figure*}

\section{Conclusion}
In this paper, we proposed the RPANet to solve the unsupervised adaptation of polyp segmentation models in the absence of source data.
The key idea of our method is to endow the target models with the capability of distinguishing foregrounds and backgrounds via self-supervision in a coarse-to-fine manner.
The proposed RPANet instantiates this objective with the incorporation of two elaborate modules, \emph{i.e.,} FCL and CCPL. FCL and CCPL learn region-level and pixel-level discriminative representations via supervised contrastive learning and confidence-calibrated pseudo-label refinement, respectively.
Experiments on three cross-domain poly segmentation tasks verified the effectiveness of our method, revealing the possibility of SFDA for real-world medical applications.

\bibliographystyle{splncs04.bst}

\bibliography{reference.bib}

\end{document}